\documentclass[letterpaper, 10 pt, conference]{ieeeconf}  
\IEEEoverridecommandlockouts
\usepackage{graphicx}
\usepackage{epstopdf}
\usepackage{amsmath}
\usepackage{amssymb}
\usepackage{subfigure}
\usepackage{multirow}
\usepackage{pbox}
\usepackage{cite}
\usepackage{algorithm}
\usepackage[noend]{algpseudocode}
\usepackage{booktabs}
\usepackage{wrapfig}
\usepackage{lscape}
\usepackage{bm}
\usepackage{url}
\usepackage{esvect}
\usepackage{xcolor,soul}
\usepackage{txfonts}
\usepackage[normalem]{ulem}
\usepackage{tikz}
\usetikzlibrary{matrix,calc}

\DeclareSymbolFont{extraup}{U}{zavm}{m}{n}
\DeclareMathSymbol{\varheart}{\mathalpha}{extraup}{86}
\DeclareMathSymbol{\vardiamond}{\mathalpha}{extraup}{87}

\algnewcommand\algorithmicforeach{\textbf{for each}}
\algdef{S}[FOR]{ForEach}[1]{\algorithmicforeach\ #1\ \algorithmicdo}

\newtheorem{definition}{Definition}

\renewcommand{\baselinestretch}{0.94}

\begin{document}

\title{\LARGE \bf Epistemic Planning for Heterogeneous Robotic Systems}
\author{Lauren Bramblett and Nicola Bezzo%
\thanks{Lauren Bramblett and Nicola Bezzo are with the Departments of Systems and Information Engineering and Electrical and Computer Engineering, University of Virginia, Charlottesville, VA 22904, USA. 
Email: {\tt \{qbr5kx, nb6be\}@virginia.edu}}}

\maketitle

\begin{abstract} 
In applications such as search and rescue or disaster relief, heterogeneous multi-robot systems (MRS) can provide significant advantages for complex objectives that require a suite of capabilities. However, within these application spaces, communication is often unreliable, causing inefficiencies or outright failures to arise in most MRS algorithms. Many researchers tackle this problem by requiring all robots to either maintain communication using proximity constraints or assuming that all robots will execute a predetermined plan over long periods of disconnection. The latter method allows for higher levels of efficiency in a MRS, but failures and environmental uncertainties can have cascading effects across the system, especially when a mission objective is complex or time-sensitive. To solve this, we propose an epistemic planning framework that allows robots to reason about the system state, leverage heterogeneous system makeups, and optimize information dissemination to disconnected neighbors. Dynamic epistemic logic formalizes the propagation of belief states, and epistemic task allocation and gossip is accomplished via a mixed integer program using the belief states for utility predictions and planning. The proposed framework is validated using simulations and experiments with heterogeneous vehicles.

\end{abstract}

\section{Introduction}
Heterogeneous multi-robot system deployment offers a variety of advantages including improved versatility, scalability, and adaptability over homogeneous systems. As robotic technology has advanced over the last few decades making robots smaller, more capable, and affordable, demand for multi-robot research has grown. Appropriate coordination of these heterogeneous systems can improve the effectiveness of safety critical missions such as surveillance, exploration, and rescue operations by incorporating the capabilities of each robot. However, the complexity of the solution for a heterogeneous system can exponentially expand over long periods of disconnectivity, especially in uncertain environments.

As humans, if the local plan must change at run-time, we are able to handle communication limitations by reasoning about the progress of other actors while empathizing with what other actors might believe about us. In our previous work \cite{Bramblett2023Epi}, we formalized a dynamical framework for logical planning using \textit{epistemic planning}, considering the knowledge and beliefs of the MRS. This method allows a distributed system to reason iteratively about the location of other robots in the system.
  
In this paper, we build on our previous work \cite{Bramblett2023Epi} and consider fault and disturbance tolerant task allocation for heterogeneous robotic systems under communication constraints. 
\begin{figure}
    \includegraphics[width = 0.48\textwidth]{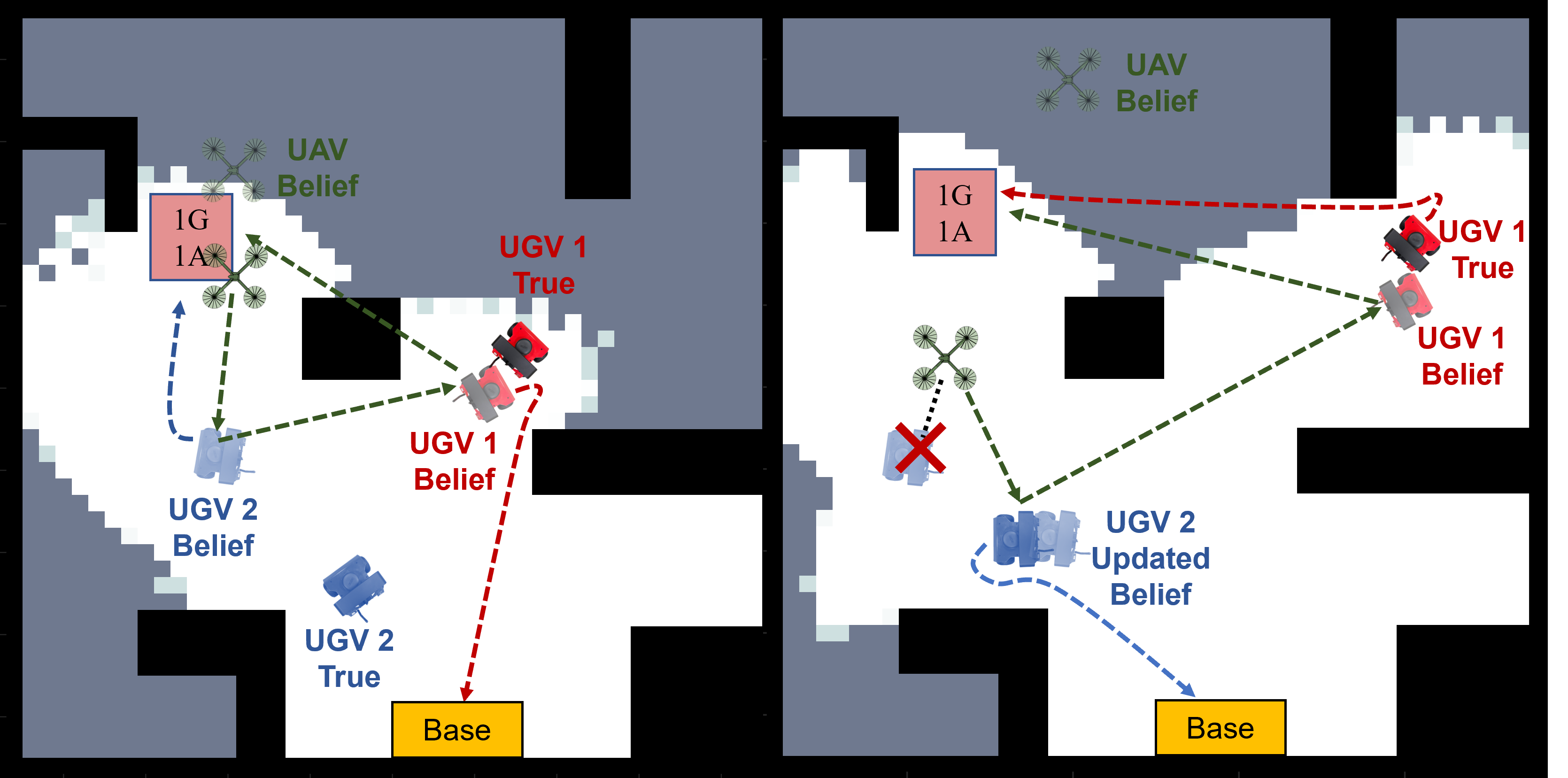}
    \vspace{-8pt}
    \caption{Pictorial depiction of the problem. The proposed framework enables a robot to reason from other agents' perspectives as it experiences a behavior change or observes that another robot is not where expected.} 
    \label{fig:introPic}
    \vspace{-20pt}
\end{figure}
We note that calculating a distributed plan for coverage while accounting for any combination of robot system failures, changes in the environment, or deviations is intractable. Instead, we propose a reasoning framework using dynamic epistemic logic where each robot $i$ propagates \textit{belief} states that represent where the robot believes other robots could be and \textit{empathy} states that represent what other robots might believe about robot $i$. Using epistemic planning, each robot can reason about a distributed strategy given local observations to complete any combination of mission objectives such as communicating with another robot, exploring an environment, or performing tasks.

For example, consider Fig.~\ref{fig:introPic} where two unmanned ground vehicles (UGVs) and one unmanned aerial vehicle (UAV) are exploring an environment and may discover tasks at undisclosed locations.  During disconnection, the UAV maintains a set of possible (belief) states for UGV 1 and UGV 2 and also a set of (empathy) states that UGV 1 and UGV 2 might believe about the UAV. The UAV finds a task that requires a UGV and plans to communicate with UGV 2. After the UAV travels to UGV 2's first belief state, it finds that UGV 2 is not present. So, the UAV reasons that UGV 2 might be at the second belief state, successfully communicates, and plans to utilize UGV 1 instead of UGV 2 due to UGV 2's deprecated state, sending UGV 2 to the base. 
In this way, robots can iteratively reason using their local observations, even when the plan requires a change at runtime.

Thus, the main contribution of this work is a formal heterogeneous task assignment and communication framework using epistemic planning for complex tasks considering disturbances and uncertainties in a communication restricted environment. 
We show that the performance of our method correlates with cases where continuous connectivity is guaranteed and outperforms motion planning methods that require the robots to be constantly connected for the entire duration of the operation. 


\section{Related Work}
\label{sec:relatedwork}
Multi-robot systems have received considerable attention in recent years due to their scalability, versatility, and applicability to various application domains \cite{rizk2019cooperative,darmanin2017review,zhou2021fuel}. A well-studied multi-robot research area is environmental exploration. Authors in \cite{corah2019communication} and \cite{yang2020multi} study multi-robot exploration missions in both known and unknown environments, but rely on continuous connectivity. Some recent works consider communication limitations by restricting motion planning to maintain connectivity \cite{capelli2020connectivity}, assigning unique objectives for sub-teams of the system \cite{bartolomei2023towards}, or using intermediary radios for connectivity maintenance \cite{saboia2022achord}. 

A related area of study is multi-robot task allocation (MRTA). MRTA is a combinatorial optimization problem and involves assigning a subset of robots to optimize the completion of an objective where task objectives can have complexities such as extended time assignment, requiring multiple robots, or having precedence constraints \cite{choi2009consensus}. Authors in \cite{sakamoto2020routing} solve a variant of heterogeneous MRTA by maximizing the reward of a heterogeneous team using a self-organizing map heuristic. Few works have also included communication limitations and failures in their approach. \cite{chen2022consensus} includes multi-robot communication limitations and allocates tasks using a consensus-based bundling algorithm with connected robots. Similarly, \cite{khodayi2019distributed} accounts for communication limitations and assigns targets to individual teams and plans rendezvous with team members to reduce the uncertainty of targets over time.   Authors in \cite{al2018generation} include system failures in their multi-robot policy search, but assume that robots are able to communicate these disruptions. 

Although research in multi-robot task allocation and environmental exploration have incorporated realistic limitations, there are few applications that combine prolonged or intentional disconnection and system failures. \cite{otte2020auctions} is one such work that uses an auction allocation algorithm to assign tasks in a communication limited environment, but assumes that the number of locally connected robots is adequate to accomplish discovered tasks. Other works deal with prolonged disconnections by establishing rendezvous points \cite{gao2022meeting,Bramblett2022}; however, this method can introduce unnecessary communication and laborious backtracking. In contrast, this work applies dynamic epistemic logic (DEL) \cite{van2007dynamic}, allowing each robot in the MRS to reason and plan using its beliefs of other robots in the system while disconnected, updating its beliefs if new events are observed and routing to communicate when necessary. DEL is a formal logic that describes how beliefs and knowledge change and has recently been integrated into robotics applications. The method presented in \cite{bolander2021based} recreates the Sally-Anne psychological test for human-robot interactions. Typical DEL-based multi-robot research uses epistemic planning for game theory-based policies \cite{maubert2021concurrent}. We extend DEL use to a realistic multi-robot application to reason about the system's state considering environmental disturbances, system failures, task discovery, and partially-unknown environments. 

\section{Preliminaries}
\label{sec:preliminaries}
\subsection{Notation, Communication, \& Control}
Given a heterogeneous multi-robot system of $N_r$ robots in the set $\mathcal{A}$, we consider a system with a set of capabilities represented by $\mathcal{Q}$ where each $q_i\in\mathcal{Q}$ is the capability of the $i^\text{th}$ robot. The element $q_i$ represents a possible subset of $N_k$ capabilities.
Additionally, all initial positions of the robots are known and the MRS's connectivity graph is denoted as $G=(\mathcal{A},\mathcal{E})$ where the set $\mathcal{E} \subset \mathcal{A} \times \mathcal{A}$ represents edge connections between robots. An edge $(i,j) \in \mathcal{E}$ indicates that robots $i$ and $j$ are within communication range (i.e. connected).

We let $N_t$ signify the number of complex tasks (i.e., tasks that require solution strategies such as multi-robot tasks or tasks with precedence relations) in the set $\mathcal{T}$ located at initially unknown positions within the operating environment. The number of tasks may be known or unknown a priori. An element $\tau$ in $\mathcal{T}$ is defined by the tuple identifying the location, the required robot capabilities, and the duration of the task: $(x_\tau,y_\tau,R_\tau,\delta_\tau)$ where $R_\tau\subseteq Q$. We assume that the tasks are stationary and completed once a subset of robots navigate within a radius $r_c\in\mathbb{R}_{> 0}$ for a length of time, $\delta_\tau$. 
The robots are assigned to search for these tasks in an environment, $\mathcal{W}$, which is modeled as an occupancy grid, $\mathcal{M} \subseteq \mathbb{R}^2$. As robots observe unexplored cells $\mathcal{M}_u \subseteq \mathcal{M}$, $\mathcal{M}$ is updated using recursive Bayesian estimation, although any applicable method can be used. We also define the frontier set $\mathcal{F}\subseteq \mathcal{M}\setminus \mathcal{M}_u$ as the set of explored cells adjacent to unknown cells and $\mathcal{O}\subseteq\mathcal{M}$ as the set of known obstacles assuming that the environment is partially-unknown.

We let $\bm{x}_i(t)$ denote robot $i$’s state variable that evolves according to general linear or nonlinear dynamics:
\begin{equation}
   \bm{x}_i(t+1) = \mathbf{g}(\bm{x}_i(t),\bm{u}_i(t),\bm{\nu}_i(t))
   \label{eq:NLDynamics}
\end{equation}
where $\bm{u}_i\in\mathbb{R}^{d_u}$ and the variable $\bm{\nu}_i\in\mathbb{R}^{d_\nu}$ denotes the control input and zero-mean Gaussian process uncertainty at discrete time $t$. The tuple $\Omega_i(t) = (\bm{x}_i(t),q_i,M_i,\sigma_i)$ of robot $i$ is referred to as robot $i$'s disposition and is defined by robot $i$'s state, capability $q_i$, local occupancy map $M_i$, and status $\sigma_i$.
\textit{Status} 
is defined as a robot's current objective such as covering the environment, communicating, or performing a task. We let a robot $i$'s status be a proposition that represents the objective a robot is executing. 

\subsection{Epistemic Logic}
\label{sec:epiLog}
In this work, distributed knowledge and reasoning for robots in the system is modeled using epistemic logic \cite{rendsvig2019epistemic}. An epistemic state is formally defined as follows:
\begin{definition}
An epistemic state is described for a finite set of atomic propositions, $AP$,
by the tuple $s = (W,R_{i},V,W_d)$ where
\begin{itemize}
\item $W$ is a non-empty, finite set of possible worlds
\item $R_i\subseteq W\times W$ is an accessibility relation for robot $i$
\item $V\rightarrow 2^{AP}$ is a valuation function.
\item $W_d\subseteq W$ is the set of designated worlds from which all worlds in $W$ are reachable.
\end{itemize}
\end{definition}
The logical formula $vR_i w$ is interpreted as ``though the true world is $w$ robot $i\in\mathcal{A}$ believes the world is $v$." The variable $s$ represents an epistemic state and we set the initial epistemic state to $s_0 = (W,R,V,w_0)$. If $W_d=\{w_0\}$, $s_0$ is the global epistemic state. The world, $w$, signifies a set of true propositions which in our application is the status of each robot $w=\{\sigma_i\forall i \in\mathcal{A}\}$. The worlds that the system can be in are described by the combinations of all possible statuses of each robot in the multi-robot system.

We define beliefs to be the estimated state of a robot with a pre-defined set of failures (i.e., faults or disturbances). Each robot predicts the future states of a set of beliefs for all robots in the system. The set $\mathcal{P} = \{\mathcal{P}_1 , \dots , \mathcal{P}_{N_a} \}$ holds the distributed beliefs of all robots. An element in $\mathcal{P}_i$ represents possible states from a robot $i$'s perspective of robots $j \in \mathcal{A}$. For this application, the epistemic language, $\mathcal{L}(\Psi,\mathcal{P},\mathcal{A}$) is obtained as follows in Backus-Naur form \cite{knuth1964backus}:
\begin{equation*}
    \phi \Coloneqq H(\omega) \ | \ \phi\land\phi \ | \ \neg\phi \ | \ K_i\phi \ | \ B_i\phi  
\end{equation*}
where  $i,j\in\mathcal{A}$. $H\in\Psi$ with $\Psi$ signifying a set of functions that describe the system state. $\omega$ generally denotes function arguments while $\neg\phi$ and $\phi\land\phi$ are propositions that can be negated and form logical conjunctions. $K_i\phi$ and $B_i\phi$ are interpreted as ``robot $i$ knows $\phi$" and ``robot $i$ believes $\phi$", respectively.

Dynamic epistemic logic is expanded from epistemic logic through action models \cite{rendsvig2019epistemic}. These models affect a robot's perception of an event and influence its set of reachable worlds. 
We simplify the notation of the action model by referring to actions in plain language. In this paper, we describe a robot's main actions or action library, $A$, as: {\em perceive} a robot, belief, or task and {\em announce} a proposition or system state. Further, we express the epistemic product model as $s\otimes i:a = (W',R_i',V',W_d')$ where $i:a$ specifies that an action $a\in A$ has been executed by robot $i$. 

\section{Problem Formulation} \label{sec:probform}
In this paper, we consider a scenario in which a heterogeneous MRS must coordinate in a decentralized fashion to efficiently search for $N_t$ complex tasks at unknown locations in a communication restricted, partially-unknown environment with the potential for system failures or disturbances. There are two main challenges that arise from this scenario: 1) how to encourage efficient exploration of a partially-unknown environment with limited communication and 2) how to efficiently calculate a plan to accomplish mission objectives, accounting for the potential of failures/disturbances that lead to different abilities of robots in the system during disconnection.
Formally, we define this problem as follows:

\textit{Problem 4.1 (Epistemic heterogeneous task allocation):} 
Given a heterogeneous, multi-robot system made of $N_r$ robots having $N_k$ different capabilities, find a distributed, sequential policy, $\pi$, to enable the multi-robot system to quickly perform cooperative search with limited communication, in a cluttered environment $\mathcal{W}$, containing $N_t$ tasks at unknown locations, with each task $\tau$ requiring a subset of the available system capabilities, $R_\tau\subseteq\mathcal{Q}$. The policy should minimize mission time while enabling cooperative behavior if a robot's abilities change while disconnected.

\section{Approach}
\label{sec:approach}
\begin{figure}[t]
\vspace{-2pt}
    \includegraphics[width = 0.48\textwidth]{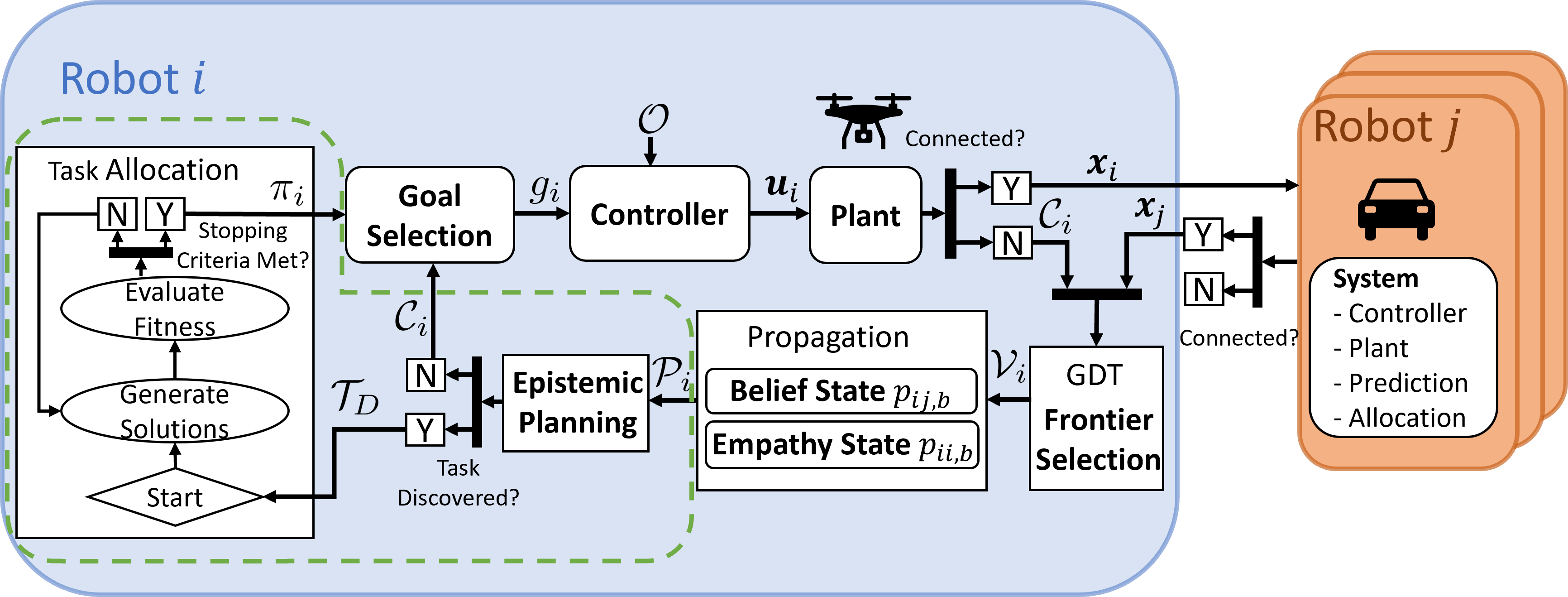}
    \vspace{-5pt}
    \caption{Diagram of the proposed approach. The contributions of this paper are within the green box.}
    \label{fig:mainFrame}
    \vspace{-19pt}
\end{figure}

Our proposed framework propagates belief and empathy states to inform heterogeneous exploration and goal assignment, considering task discovery and system failures in a partially-unknown environment. For ease of discussion let us consider two robots $i$ and $j$. From robot $i$'s perspective, a {\em{belief state}}, $p_{ij,b}\in\mathcal{P}_i$, represents a possible state of a robot $j$ and an {\em{empathy state}}, $p_{ii,b}\in\mathcal{P}_i$, describes robot $i$'s belief of robot $j$'s belief about robot $i$'s state. With this knowledge, robot $i$ predicts and tracks empathy states to ensure that a robot $j$ holds one true belief of the state of robot $i$ if covering the environment (i.e. robot $i$ will plan its motion according to at least one empathy particle). Thus, robot $i$'s empathy states are equivalent to robot $j$'s belief states for robot $i$. The diagram in Fig.~\ref{fig:mainFrame} summarizes this architecture.

Initially, as shown in Fig.~\ref{fig:mainFrame}, robot $i$ assesses if communication is successful and, if it is, frontiers are partitioned in a distributed manner and each robot calculates their respective frontiers using a generalized Dirichlect tessellation (GDT) \cite{ash1986generalized}. If the robots disconnect, the common belief set, $\mathcal{C}_i$, acts as the state in the GDT for any robot $j \in \mathcal{A}$ from $i$'s perspective that is not connected. Each robot plans its path to a frontier point using a smooth A$^*$ \cite{mueggler2014aerial} and local control for path following using an artificial potential field.

While searching the environment, the robot system's main purpose is to execute tasks. Upon discovery, a policy, $\pi_i$, is established for each robot using a non-linear integer program. If robot $i$ discovers a task that requires robot $j$, the robot includes the task of communicating with robot $j$ (gossiping) using the set of beliefs for robot $j$, $\{p_{ij,b}, \forall b\in\mathcal{B}\}$ and updates its policy, $\pi_i$. 

In the following sections, we will describe the principal elements in our framework including: i) belief and empathy propagation, ii) epistemic coverage, iii) epistemic planning, and vi) task allocation and gossiping protocol.

\subsection{Belief \& Empathy Propagation}
\label{sec:bepropagation}
In our epistemic framework, each robot propagates belief states for all robots in the multi-robot system. This allows a robot $i$ to plan according to its beliefs about other robots and empathize with what other robots expect robot $i$ to accomplish while disconnected. To constrain system uncertainties while disconnected to a finite set of possibilities, we define a finite set of particles, $\mathcal{P}_i$, to represent these belief and empathy states from the perspective of the $i^{th}$ robot: 
\begin{equation}
    \mathcal{P}_i=\{p_{ij,b} \ \forall j\in\mathcal{A},\forall b\in\mathcal{B}\}.
\end{equation}
The $i^{th}$ robot defines its empathy particles as $\mathcal{P}^e_{i} = \{p_{ii,b} \ \forall b\in \mathcal{B}\}$ and its belief particles about other robots as $\mathcal{P}^r_{i} = \{p_{ij,b}  \ \forall j\in \mathcal{A},\forall b\in \mathcal{B}\}$. For each robot $j \in \mathcal{A}$, the robot $i$ orders its belief and empathy particles $1$ through $N_b$ according to the probability of occurrence (that is, from largest to smallest). The order is initialized prior to deployment and each robot $i$ initially tracks its first empathy particles.

If disconnected, a robot $i$ propagates beliefs from the last globally communicated state between robot $i$ and robot $j$. We define this set of particles as $\mathcal{C}_i \subseteq \mathcal{P}_i$ and refer to it as robot $i$'s \textit{common belief} set. The motion plan for all particles is computed using the common belief set, $\mathcal{C}_i = \{c_{ij} \ \forall j\in\mathcal{A}\}$ and the 
dynamics defined in \eqref{eq:NLDynamics}. 

The goal selection for each robot and particle is dependent on the believed status of a robot. Within this paper, the main statuses for a robot are: \textit{exploring}, \textit{gossiping}, 
 or \textit{performing a task}, noting that these statuses are predefined and mission dependent. Given that all robots follow an empathy particle during exploration, we next present our strategy to propagate these states while disconnected.

\subsection{Epistemic Coverage Assignments}
As noted in the related work, most distributed algorithms rely on a fully connected system. In contrast, we utilize a partitioning and coverage mechanism using the common belief set, $\mathcal{C}$, for cooperative robots given a partially-unknown environment while disconnected. 

To begin, an $i^\text{th}$ robot updates its local map and the estimated coverage from robot $j$ at the location described by the $j^\text{th}$ particle in the set $\mathcal{C}_i$ using recursive Bayesian estimation. Using this common belief map, an $i^\text{th}$ robot determines its frontier set, $\mathcal{F}_i$, by assessing which explored cells are adjacent to unknown cells. Additionally, the optimal partition of $\mathcal{F}_i$ is the tessellation $\mathcal{V}_i(\mathcal{C}_i) = \{\mathcal{V}_{i1},\mathcal{V}_{i2},\dots,\mathcal{V}_{iN_r}\}$ generated by common belief particles in $\mathcal{C}_i$ denoted as the points $(c_{i1},c_{i2},\dots,c_{iN_r})$ and weighted by a constant factor $\psi_j$ based on a $j^\text{th}$ robot's capability:
\begin{equation}
\mathcal{V}_{ij} = \{f\in\mathcal{F}_i | \ \  \psi_j ||f-c_{ij}||\leq \psi_k ||f-c_{ik}||, \forall k\neq j\}.\label{eq:vornoi}
\end{equation}
With each robot's frontier assignment given in \eqref{eq:vornoi}, we can determine the utility of each frontier point. The utility of a frontier point is user-defined (e.g. travel time, distance to other robots) and includes a penalty for frontier points outside of a particles' partition. The utility of each frontier point is defined as:
\begin{equation}
\upsilon_{ij,z} = \begin{cases} 
      h(f_z,c_{ij}) + \Delta & f_z\not\in\mathcal{V}_{ij}\\
      h(f_z,c_{ij})  & f_z\in\mathcal{V}_{ij}      
   \end{cases}
   \label{eq:frontierPoint}
\end{equation}
where $\Delta$ is a user-defined penalty for frontier points outside of a robots' partition and $h(\cdot)$ is the utility function for assigning $c_{ij}$ to $f\in\mathcal{F}_i$. Subsequently, the frontier point that minimizes the utility from \eqref{eq:frontierPoint} is defined as
\begin{equation}
    z^* = \underset{z}{\text{argmin}} \ \upsilon_{ij,z}
\end{equation}
and
\begin{equation}
    g^c_{ij} = f_{z^*}.
\end{equation}
The variable $g^c_{ij}$ is the frontier point goal for the common belief particle, $c_{ij}$, which encourages the common belief to propagate to unique and uncovered portions of the environment. If a particle's status is {\em exploring}, it also shares the same goal as its respective common belief particle: $g_{ij,b} = g^c_{ij} \ \forall j\in\mathcal{A},\forall b\in\mathcal{B}$. Otherwise, the goal for each particle depends on the particle's status, such as going to a task or communicating with another robot. Once all frontier points are believed to have been visited, the robot $i$'s particles converge to a common meeting place using the mean location of the  particles in the set $\mathcal{C}_i$: 
\begin{equation}
    \sum_{c_{ij}\in \mathcal{C}_i} c_{ij}/|\mathcal{C}_i|
    \label{eq:commonMeeting}
\end{equation} where $|\cdot|$ signifies the cardinality of a set.

If a robot experiences a failure, we assume that each robot $i$ is capable of computing the set of empathy states that are suitable to track. For robot $i$, we denote this set as $\mathcal{P}^t_{i}\subseteq\mathcal{P}^e_{i}$. The robot chooses to track the particle in $\mathcal{P}^t_{i}$ with the highest likelihood 
If all robots are within communication range, the first particle becomes the robot's current state and subsequent particles are propagated based on the updated vehicle state.

\begin{figure}[h]
\vspace{-0pt}
    \centering
    \subfigure[]{\includegraphics[width=0.15\textwidth]{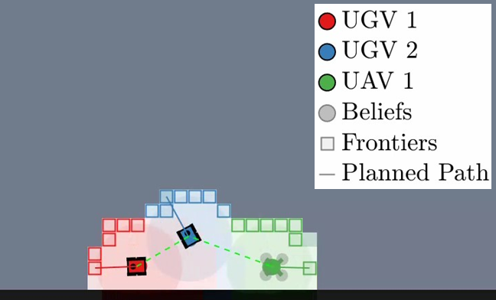}\label{fig:propFig0}}
    \subfigure[]{\includegraphics[width = 0.15\textwidth]{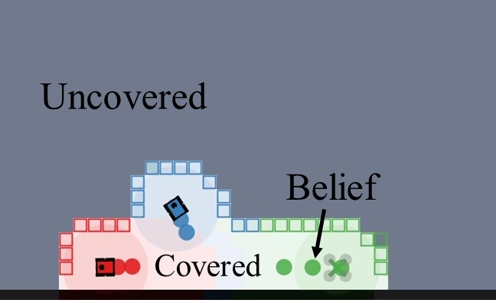}\label{fig:propFig1}}
    \subfigure[]{\includegraphics[width = 0.15\textwidth]{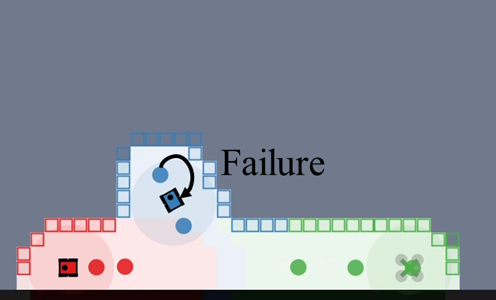}\label{fig:propFig2}}
    \vspace{-7pt}
    \caption{Depiction of particle propagation over time for three vehicles at initialization (a), disconnection (b), and for a failure (c).}
    \label{fig:exampleProp}
    \vspace{-5pt}
\end{figure}

Fig.~\ref{fig:exampleProp} shows an example of particles propagating in a partially-unknown environment for three vehicles (2 UGVs and 1 UAV). In Fig.~\ref{fig:propFig0}, the robots begin within communication range and establish goals along the frontier using \eqref{eq:frontierPoint}. In Fig.~\ref{fig:propFig1}, the robots disconnect moving toward their respective coverage goals and establish belief states. The number of belief states is finite so that if robot $i$ experiences a failure and can no longer track the common belief, robot $i$ will track the next particle that is more likely to empathize with other $j$ robots. Similarly, if robot $j$ needs to communicate with robot $i$ it will begin by checking the common belief and iterate through the next most probable particles to find robot $i$. 
Fig.~\ref{fig:propFig2} shows UGV 2 experiencing a failure and tracking its second particle. The covered area is shaded by the robot color that accomplished coverage, and the plotted frontier points are the frontier points from each belief states' perspective, dynamically allocated using \eqref{eq:vornoi}.

\subsection{Epistemic Modeling}
As a robot explores the environment, new information may become available while disconnected from other robots. We use epistemic planning to account for each robot's imperfect knowledge of the system given the communication restrictions in our application. Though \cite{maubert2019reachability} states that multi-player games 
with imperfect information are undecidable, we use epistemic planning to reduce the computational complexity of the problem and complete all tasks in most disconnected scenarios.

When robots communicate, rational belief updates must occur to ensure that the common belief is still retained. In this work, we establish three cases where updates occur: i) when connected to all robots, ii) when expecting to connect with another robot, and iii) when the robot discovers a task. From the established semantics in Sec.~\ref{sec:epiLog}, we introduce our action library $A$ that can transform the epistemic state. We let $A=\{perceive(\phi),announce(\phi)\}$. The action $perceive$ occurs when a robot perceives a generic proposition $\phi$ in the environment, such as a task, particle, or robot,  noting that perception can occur through local communication to another robot or by direct perception. The action $announce$ is when a robot communicates with its locally connected team.  The set $\Psi=\{present,track\}$ is functionally interpreted for $K_i \, present(\tau)$ as robot $i$ knows the location, required robots, the duration of the task $\tau$ and for $K_i \, track(p_{ij,b})$ as robot $i$ knows that robot $j$ is tracking the belief particle $b$.

First, we address the global belief update when all robots are connected. We assume that because robots are cooperative, all belief updates are accepted and do not become outdated unless an event occurs in the environment such as discovering a task or system failures or disturbances. However, a robot may not know when/if the information of the system becomes outdated when disconnected. We formulate the logic for this framework using a series of worlds, $w_t$, which is the set of propositions of each robot's status, $\sigma^t_{i} \ \forall i\in\mathcal{A}$. Additionally, there exists one true world, $w^*_t$, at time $t$ and it only exists if
\begin{equation}
    w^*_tR_i w^*_{t}, \ \forall i\in \mathcal{A}.
\end{equation}
In order for all robots to know with certainty the true world, all robots' statuses, $\sigma_t^i, \ \forall i\in\mathcal{A}$ 
must be common knowledge and announced such that the epistemic state from robot $i$'s perspective at time $t$ is:
\begin{equation}
    s^i_{t-1}\otimes announce(\Omega) = s^i_t\models K_i \sigma^i_{t} \bigwedge_{j\in\mathcal{A}} K_i K_j \sigma^j_{t} \ \forall i\in\mathcal{A}.
    \label{eq:publicAnnounce}
\end{equation}
where $announce(\Omega)$ is an action symbolizing the announcement of all robots' dispositions. 

The common belief particles are updated from the announcement of all states to the multi-robot system such that
\begin{equation}
    c_{ij}\gets \Omega_j, \ \forall (i,j)\in \mathcal{A}^2.
\end{equation}

Since the common belief is updated to the world $w_t$ shared according to \eqref{eq:publicAnnounce}, the particles in this set are propagated according to the status propositions of each robot. For example, in a two-robot team, if robot 1 communicates with robot 2 that it has found a task and will complete this task, robot 1 and robot 2 would propagate a common belief particle that moves robot 1 to execute the task before continuing to cover the environment.

Local beliefs are updated differently since common belief particles are still propagated, but a robot may act on new information in the environment. There are two actions that cause a robot to change its belief of the world. First, if robot $i$ perceives that robot $j$ is not at its believed location:
\begin{equation}
s^i_{t-1}\otimes i:perceive(\neg track(p_{ij,b})) = s_t\models K_i \neg\sigma^j_{t-1}.
\end{equation}
This causes robot $i$ to update its belief of the world and robot $j$ such that
\begin{equation}
s^i_{t-1}\otimes i:perceive(\neg track(p_{ij,b})) = s_t\models B_i track(p_{ij,b+1}).
\end{equation}
Second, if a robot perceives a task, epistemic state updates occur as follows:
\begin{equation}
    s^i_{t-1}\otimes i:perceive(present(\tau)) = s_t\models K_i\sigma_t^i.
\end{equation}
In this way, the knowledge of disconnected robots is not affected, nor does robot $i$ update its belief that a disconnected robot would know the updated information.

With our epistemic states and actions defined, we now describe how these concepts can be used for planning. A planning task for a robot $i$ is defined by the tuple $\Pi=(s_t^i,A,\gamma)$ where $\gamma$ is a goal formula. In plain language, the goal formula is to complete all tasks in the environment. The execution of $\pi$ is defined as a maximal sequence that satisfies the global formula $\gamma$. 

Generally, each robot uses the belief updates defined in this section as inputs to the optimal task assignment presented in the following section. For a robot $i$, the optimized task allocation and policy for each robot depends on the believed disposition of all vehicles, modeled by the epistemic state $s_t^i$. 

\subsection{Epistemic Task Allocation \& Gossip Protocol}
Lastly, this framework requires a method for using the epistemic planning task, $\Pi$, and establishing an execution policy, $\pi$ that satisfies the mission objective. We consider the solution to a planning task as a joint policy so that each robot is responsible for subtasks within the mission objective. The solution to such a policy is solved using the below nonlinear integer program where the utility, $u(\cdot)$, is maximized. 
\begin{align}
\max \ \ & \sum_{i\in\mathcal{A}}\sum_{\tau\in\mathcal{T}} u_{i\tau}(t_{i\tau}(\bm{b}_i(\bm{y}_i),s^i_t))y_{i\tau}\label{eq:objForTA}\\
\text{s.t.} \ & \sum_{i\in\mathcal{A}}\sum_{\tau\in\mathcal{T}}y_{i\tau}\mathbf{I}^k_i\geq r^k_{\tau}, \ \ \forall \tau\in\mathcal{T}, \ \forall k\in\{1,\dots,N_k\}\nonumber\\
&t_{i\tau}(\bm{b}_i(\bm{y}_i),s_t^i)\geq t_{i\varsigma}(\bm{b}_i(\bm{y}_i),s_i^t)+\delta_{\varsigma\tau} \ \forall (\varsigma,\tau)\in S_i\label{eq:NLPconstraints}\\
&t_{i\tau}(\bm{b}_i(\bm{y}_i),s_t^i)\geq 0 \ \forall \tau\in \mathcal{T}\nonumber\\
&y_{i\tau}\in\{0,1\}\nonumber
\end{align}
where $y_{i\tau} = 1$ if robot $i$ is assigned to task $\tau$ and $\bm{y}_i=\{y_{i1},\dots,y_{iN_t}\}$. $\mathbf{I}_i^k$ is an indicator function for the robot $i$ that has the $k^\text{th}$ capability. The arrival time for the $i^\text{th}$ robot is a unique function, $t_{i\tau}$, that accounts for the arrival time of $r_\tau^k$ necessary robots with capability $k$ for task $\tau$. Additionally, the function $t_{i\tau}$ considers the current belief using the epistemic state $s_t^i$. The variable $\delta_{\varsigma\tau}$ is the duration between tasks $\varsigma$ and $\tau$. The order of tasks is represented by a directed graph, $S_i$, created by the order of robot $i$'s path, $\bm{b}_i$, where an edge $(\varsigma,\tau)\in S_i$ is indicates 
that task $\varsigma$ must be performed before task $\tau$.

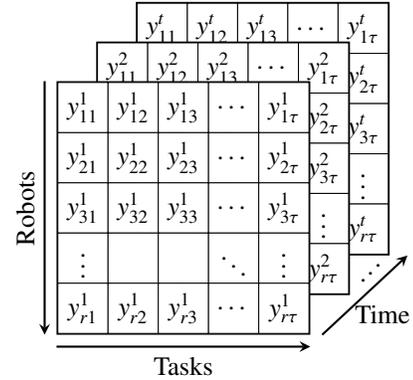
\begin{figure}[H]
\vspace{-5pt}
\centering
\begin{tikzpicture}[auto matrix/.style={matrix of nodes,
  draw,thick,inner sep=0pt,
  nodes in empty cells,column sep=-0.2pt,row sep=-0.2pt,
  cells={nodes={minimum width=1.9em,minimum height=1.9em,
   draw,very thin,anchor=center,fill=white}}}]
 \matrix[auto matrix=x, xshift=3em,yshift=3em](matz){
 $y^t_{11}$ & $y^t_{12}$ & $y^t_{13}$ & $\dots$ & $y^t_{1\tau}$\\
 $y^t_{21}$ & $y^t_{22}$ & $y^t_{23}$ & $\dots$ & $y^t_{2\tau}$\\
 $y^t_{31}$ & $y^t_{32}$ & $y^t_{33}$ & $\dots$ & $y^t_{3\tau}$\\
 $\vdots$ & & & $\ddots$ & $\vdots$ \\
  $y^t_{r1}$ & $y^t_{r2}$ & $y^t_{r3}$ & $\dots$ & $y^t_{r\tau}$\\
 };
 \matrix[auto matrix=x,xshift=1.5em,yshift=1.5em](maty){
 $y^2_{11}$ & $y^2_{12}$ & $y^2_{13}$ & $\dots$ & $y^2_{1\tau}$\\
 $y^2_{21}$ & $y^2_{22}$ & $y^2_{23}$ & $\dots$ & $y^2_{2\tau}$\\
 $y^2_{31}$ & $y^2_{32}$ & $y^2_{33}$ & $\dots$ & $y^2_{3\tau}$\\
 $\vdots$ & & & $\ddots$ & $\vdots$ \\
  $y^2_{r1}$ & $y^2_{r2}$ & $y^2_{r3}$ & $\dots$ & $y^2_{r\tau}$\\
 };
 \matrix[auto matrix=x](matx){
 $y^1_{11}$ & $y^1_{12}$ & $y^1_{13}$ & $\dots$ & $y^1_{1\tau}$\\
 $y^1_{21}$ & $y^1_{22}$ & $y^1_{23}$ & $\dots$ & $y^1_{2\tau}$\\
 $y^1_{31}$ & $y^1_{32}$ & $y^1_{33}$ & $\dots$ & $y^1_{3\tau}$\\
 $\vdots$ & & & $\ddots$ & $\vdots$ \\
  $y^1_{r1}$ & $y^1_{r2}$ & $y^1_{r3}$ & $\dots$ & $y^1_{r\tau}$\\
 };
 
 \draw[thick,-stealth] ([xshift=1ex]matx.south east) -- ([xshift=1.5ex]matz.south east)
  node[midway,below,xshift=1.5ex] {Time};
 \draw[thick,-stealth] ([yshift=-1ex]matx.south west) -- 
  ([yshift=-1ex]matx.south east) node[midway,below] {Tasks};
 \draw[thick,-stealth] ([xshift=-1ex]matx.north west)
   -- ([xshift=-1ex]matx.south west) node[midway,above,rotate=90] {Robots};
   \path (maty.south west)--(matz.south east) node[below,rotate=45,xshift=-2ex,yshift=1ex]{\ldots};
\end{tikzpicture}
\vspace{-10pt}
\caption{Pictorial representation of binary decision matrix.}\label{fig:matrixRep}
\vspace{-5pt}
\end{figure}

We use the matrix representation shown in Fig.~\ref{fig:matrixRep} to represent the solution space from \eqref{eq:objForTA}. We refer to the time of a task as an epoch in which any robot is available to perform an additional task. Using this representation, we can check precedence constraints such as gossiping to a robot before assigning it tasks to accomplish or simultaneous tasks requiring multiple robots at the task location (e.g. a UGV opening a door for a UAV to fly through). All matrix representations for one iteration of the task allocation problem are of static depth along the time axis and initialized to be the number of tasks possible to complete (i.e. gossiping tasks plus discovered tasks).

Though any method for solving a constrained nonlinear assignment problem can be used, we apply a genetic algorithm (GA) due to the sparsity of the decision space and the binary assignment constraints. Specifically, we generate an initial feasible population for the task allocation problem and then utilize a GA to efficiently sample the decision space.

Since the assignment problem must be solved at runtime, the initial population is generated using the method in Algorithm~\ref{alg:initPop} to warm-start the GA
\begin{algorithm}[h]
\caption{Feasible Solution Generation}\label{alg:initPop}
 \hspace*{\algorithmicindent} \textbf{Input:} $\Pi = (s_t^i,A,\gamma)$ \\
 \hspace*{\algorithmicindent} \textbf{Output:} Solution representing policy, $\pi$ for $\Pi$
\begin{algorithmic}[1]
\State $v_0$ is the set of connected robots and $v_r = v_0$
\While{$\tau\in\mathcal{T}_D$ not complete}
\ForEach{$\tau\in\mathcal{T}_D$}
\If{$v_r$ has the capabilities to perform the task}
\If{$v_r\equiv\mathcal{A}$ or $\text{rand}()>$threshold}
\State $v_t \in _R v_r $ for each task capability
\State $\bm{b}_j\gets \bm{b}_j\oplus$ task $\forall j\in v_t$
\EndIf
\EndIf
\EndFor
\If{$\tau\in\mathcal{T}_D$ is complete}
\State break
\EndIf
\ForEach{$v_r\notin v_t$}
\State $v_g\in_R \mathcal{A}\setminus v_r$
\State $\bm{b}_{v_r}\gets b_{v_r}\oplus v_g$
\EndFor
\State $v_r = v_r\oplus v_g$
\EndWhile
\State $\pi \gets \bigcup_{j\in\mathcal{A}}\bm{b}_j$
\end{algorithmic}
\end{algorithm}
where $\in_R$ indicates a uniformly selected element. The operator $\oplus$ appends the antecedent set to the precedent set. The output policy $\pi$ is a sequence defined by the joint execution order $\{\bm{b}_j\ \forall j\in \mathcal{A}\}$ and represented as a sequence of epistemic states and robot action pairs, $\pi = (s_t^i,(j:a),s_{t+1}^i,\dots)$. The algorithm is run for the desired size of the initial GA population.

Additionally, the constrained optimization function in \eqref{eq:objForTA} is transformed into an unconstrained, penalty-based function such that
\begin{equation}
    val = \sum_{i\in\mathcal{A}}makespan(\bm{b}_i) + V_i\label{eq:chromVal}
\end{equation}
where $makespan(\bm{b}_i)$ is the estimated length of time for robot $i$ to complete its assigned tasks and $V_i$ is the penalty for violated constraints in \eqref{eq:NLPconstraints}. Since only policies where no constraints are violated are valid for the goal formula $\gamma$, $V_i$ must be set at a high value to ensure that the selected solution of the GA is feasible. Algorithm~\ref{alg:initPop} ensures that the initial population of solutions are all valid policies for the goal formula $\gamma$, but we use the GA to iterate our solution and attempt to achieve a higher fitness value. 
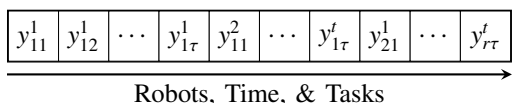
\begin{figure}[b]
\vspace{-10pt}
\centering
\begin{tikzpicture}[auto matrix/.style={matrix of nodes,
  draw,thick,inner sep=0pt,
  nodes in empty cells,column sep=-0.2pt,row sep=-0.2pt,
  cells={nodes={minimum width=1.9em,minimum height=1.9em,
   draw,very thin,anchor=center,fill=white}}}]
 \matrix[auto matrix=x](matx){
 $y^1_{11}$ & $y^1_{12}$ & $\dots$ &$y^1_{1\tau}$ & $y^2_{11}$ & $\dots$ & $y^t_{1\tau}$ & $y^1_{21}$ &  $\dots$ & $y^t_{r\tau}$\\
 };
  \draw[thick,-stealth] ([yshift=-1ex]matx.south west) -- 
  ([yshift=-1ex]matx.south east) node[midway,below] {Robots, Time, \& Tasks};
\end{tikzpicture}
\vspace{-10pt}
\caption{GA solution representation}\label{fig:chromosome}
\vspace{-0pt}
\end{figure}

The resultant solution or gene is shown generically in Fig.~\ref{fig:chromosome} where the highest fitness value is the execution policy for the robots connected locally. To achieve the highest fitness value at the lowest computational complexity, the chromosome is formatted as a single row sparse matrix. By using this representation we can utilize point mutation, one-point crossover, and roulette wheel selection \cite{chen2003genetic} such that the computational complexity of our task allocation algorithm is $O(nm^2)$ with $n$ being the number of robots and $m$ being the number of tasks.

\begin{figure}[h]
    \vspace{-5pt}
    \centering
    \subfigure[]{
    \includegraphics[width = 0.145\textwidth,trim = {0cm 0cm 0cm 0},clip]{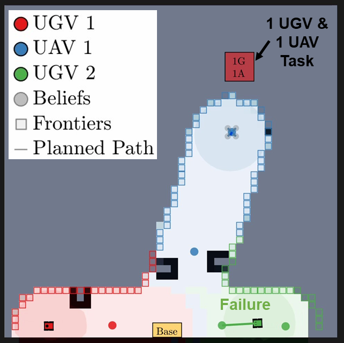}
    \label{fig:appa}
    }%
    \subfigure[]{
    \includegraphics[width = 0.145\textwidth,trim = {0cm 0cm 0cm 0},clip]{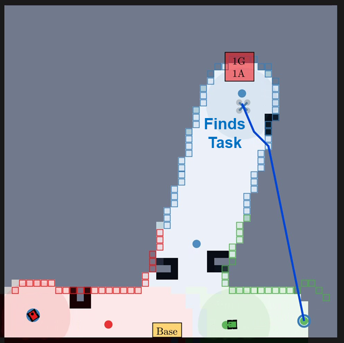}
    \label{fig:appb}
    }%
    \subfigure[]{
    \includegraphics[width = 0.145\textwidth,trim = {0cm 0cm 0cm 0},clip]{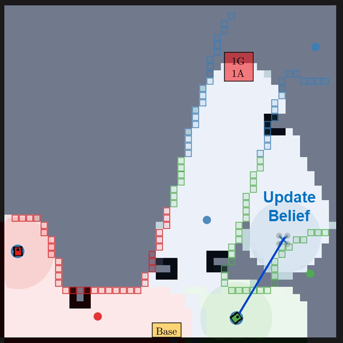}
    \label{fig:appc}
    }%
    \vspace{-5pt}
    \\
    \subfigure[]{
    \includegraphics[width = 0.145\textwidth,trim = {0cm 0cm 0cm 0},clip]{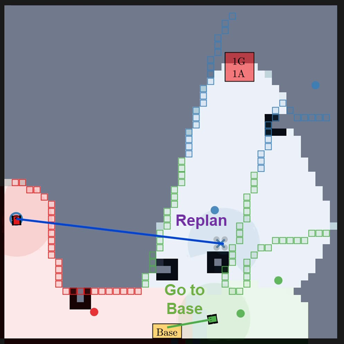}
    \label{fig:appd}
    }%
    \subfigure[]{
    \includegraphics[width = 0.145\textwidth,trim = {0cm 0cm 0cm 0},clip]{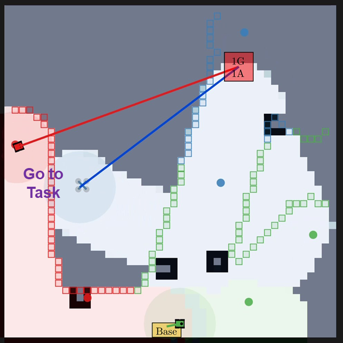}
    \label{fig:appe}
    }%
    \subfigure[]{
    \includegraphics[width = 0.145\textwidth,trim = {0cm 0cm 0cm 0},clip]{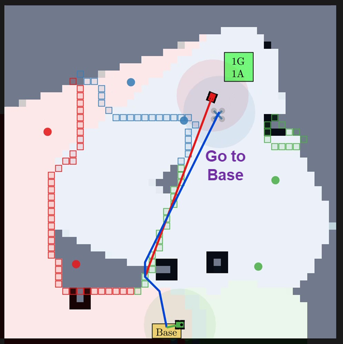}
    \label{fig:appf}
    }%
\vspace{-5pt}
\caption{Example simulation with one known task at an unknown location.}
\label{fig:approachExampFig}
\vspace{-7pt}
\end{figure}

\textbf{\textit{Example.}} To reinforce the proposed approach in the reader's mind, consider the scenario in Fig.~\ref{fig:approachExampFig} where all robots know there is one task at an unknown location (e.g., a search and rescue mission). Fig.~\ref{fig:appa} shows the robots disconnecting and beginning to propagate belief states. UGV 2 experiences a failure and moves to the second particle. In Fig.~\ref{fig:appb}, UAV 1 finds a task that requires one aerial vehicle and one ground vehicle and uses the first particle of UGV 2 in the allocation optimization. In Fig.~\ref{fig:appc}, UAV 1 observes that UGV 2 is not tracking its first particle and reallocates using UGV 2's second particle. UAV 1 is able to communicate with UGV 2 at its second belief particle and re-optimizes assignments using the new information. This sends UGV 2 back to base and assigns UGV 1 to perform the task shown in Fig.~\ref{fig:appd}. In Fig.~\ref{fig:appe} UAV 1 and UGV 1 plan to perform the task and complete the task in Fig.~\ref{fig:appf} before relocating to the base.

In general, employing this framework allows a heterogeneous MRS to propagate beliefs, reason about environmental changes, and plan according to local observations while disconnected. The output of our proposed methodology is a sequential policy that is formulated using belief states to reason about other robots location and used to accomplish any mission objectives discovered in the environment.

\section{Simulations} \label{sec:sims}
In this section, we provide results and comparisons from MATLAB simulations with our approach implemented on two, three, five, and eight robot teams. Each scenario has a randomly generated team makeup consisting of two types of vehicles, UAVs and UGVs, for each run. Simulations were performed in 15 random $50\text{m}\times 50\text{m}$ environments with 5-15 initially unknown obstacles. The robots start by assuming that the environment has no obstacles and do not know the location of the tasks. 

Each robot propagates three particles. Particles are propagated on the basis of the maximum speed of the vehicle type. A UGV may travel at a max speed of 2m/s, and a UAV may travel at 6m/s. The second and third particles travel at a linear speed decreased from the vehicle's maximum speed of 40\% and 80\%, respectively. The particles propagate according to these velocities, and the maximum communication range is 10m from the center of the robot. Within our simulations, each scenario was run with zero, one, and two failures that can happen to any robot at any time, causing the affected robot to track its second or third empathy particle.

\begin{figure}[ht!]
    \vspace{-5pt}
    \centering
    \subfigure[]{
    \includegraphics[width = 0.145\textwidth,trim = {0cm 0cm 0cm 0},clip]{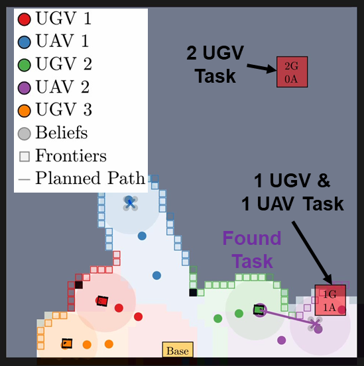}
    \label{fig:sima}
    }%
    \subfigure[]{
    \includegraphics[width = 0.145\textwidth,trim = {0cm 0cm 0cm 0},clip]{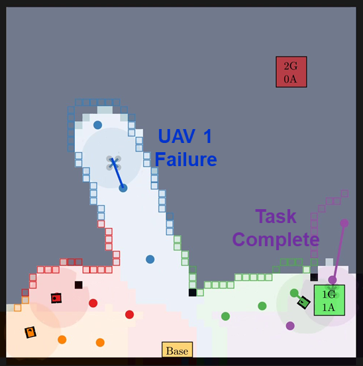}
    \label{fig:simb}
    }%
    \subfigure[]{
    \includegraphics[width = 0.145\textwidth,trim = {0cm 0cm 0cm 0},clip]{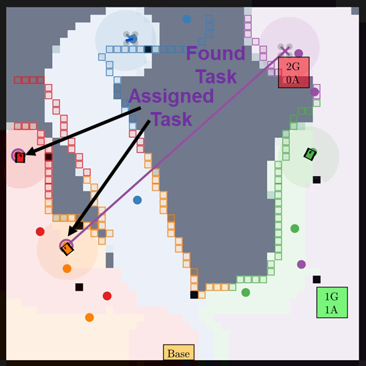}
    \label{fig:simc}
    }%
    \vspace{-5pt}
    \\
    \subfigure[]{
    \includegraphics[width = 0.145\textwidth,trim = {0cm 0cm 0cm 0},clip]{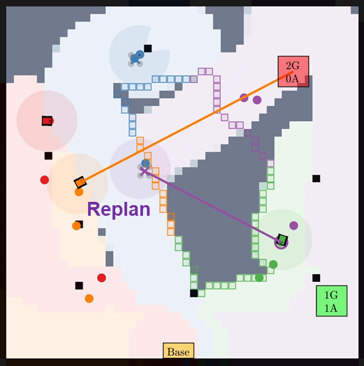}
    \label{fig:simd}
    }%
    \subfigure[]{
    \includegraphics[width = 0.145\textwidth,trim = {0cm 0cm 0cm 0},clip]{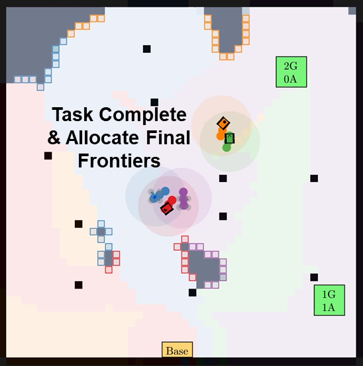}
    \label{fig:sime}
    }%
    \subfigure[]{
    \includegraphics[width = 0.145\textwidth,trim = {0cm 0cm 0cm 0},clip]{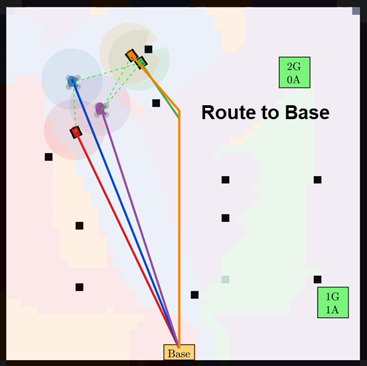}
    \label{fig:simf}
    }%
\vspace{-5pt}
\caption{Example simulation with one known task at an unknown location.}
\vspace{-5pt}
\label{fig:simExampFig}
\end{figure}

Fig.~\ref{fig:simExampFig} showcases an example of the simulation scenarios run in this section. As shown, there are two tasks in the environment, but all robots begin not knowing how many tasks or the location. Fig.~\ref{fig:sima} shows that after disconnection, one task requiring a ground and aerial vehicle is discovered by UAV 2 who communicates the task to UGV 2 and both complete the task in Fig.~\ref{fig:appb}. In Fig.~\ref{fig:appc}, UAV 2 finds a task requiring two ground vehicles. UAV 2 allocates UGV 1 and UGV 3 to the task initially, but after updating the states and replanning upon connection with UGV 1 in Fig.~\ref{fig:simd}, UAV 2 gossips to UGV 2. Both UGV 1 and UGV 2 route to the task and complete it in Fig.~\ref{fig:appe} before connecting with all robots in the system at the common meeting location and partitioning the remaining frontier. In Fig.~\ref{fig:sime}, no frontier points remain and all robots route to base.

The proposed approach is compared with two other methods. The first method, referred to as the ``flock method," applies a constant connectivity constraint, restricting all robots to travel within a 10m communication range of each other. The second ``ideal method" assumes that robots can communicate across the entire environment. Both methods use the smooth A$^*$ and APF method to control the robots towards uncovered regions and away from obstacles. In all methods, the vehicles' simulated LiDAR range is 5m and the genetic algorithm is used for task allocation. 
\begin{figure}[h]
\vspace{-0pt}
    \centering
    \includegraphics[width=0.45\textwidth]{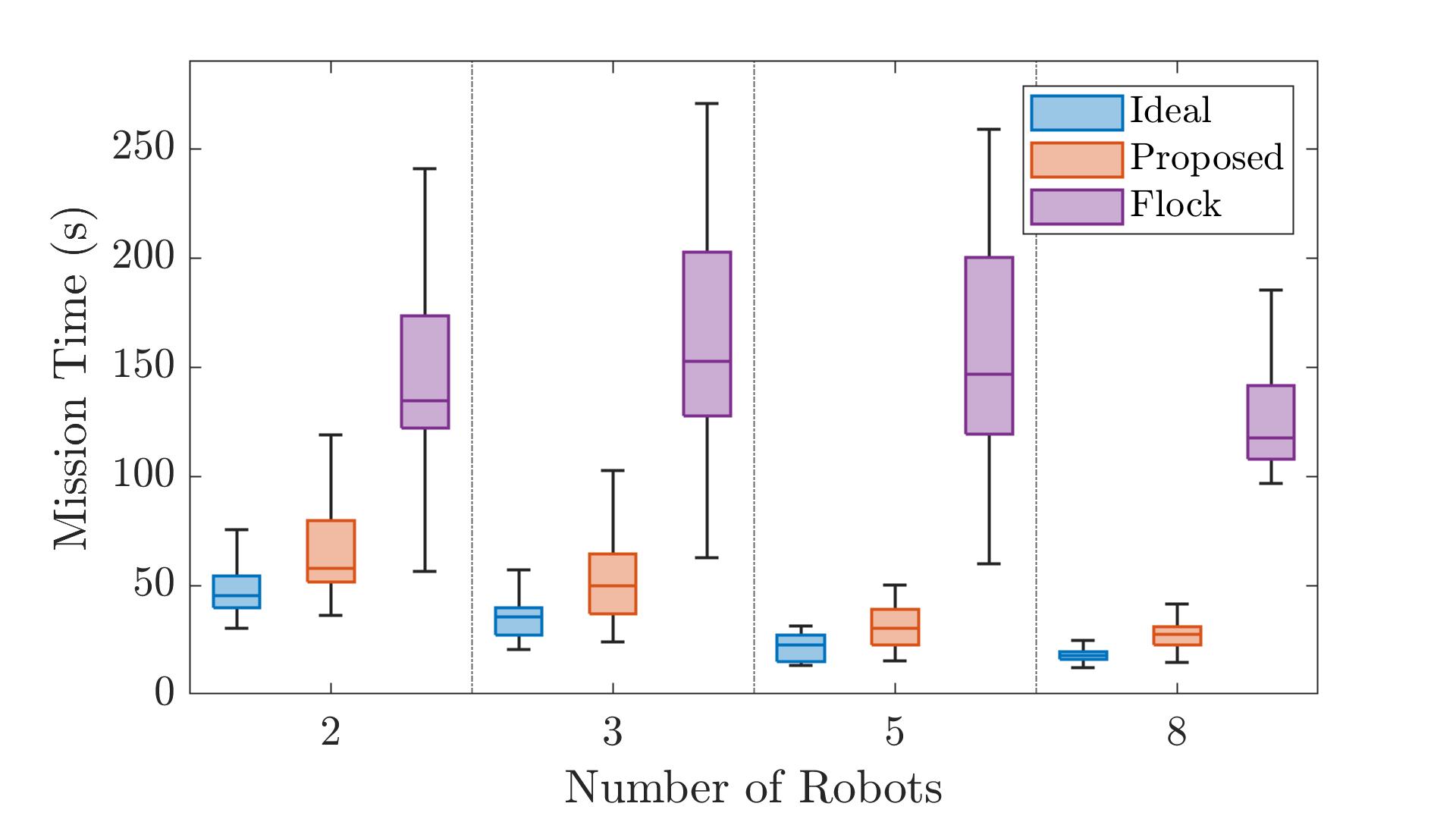}
    \vspace{-5pt}
    \caption{Comparison of the simulated scenarios.}
    \label{fig:comparisonBoxPlot}
    \vspace{-17pt}
\end{figure}

\begin{table}[b]
\vspace{-14pt}
\centering
\caption{Average mission times by simulated failure scenarios.}
\vspace{-7pt}
\label{tab:sims}
\resizebox{0.25\textwidth}{!}{%
\begin{tabular}{|l|ccc|}
\hline
                & \multicolumn{3}{c|}{\textbf{Number of Faults}}                                 \\ \hline
\textbf{Method} & \multicolumn{1}{c|}{\textbf{0}} & \multicolumn{1}{c|}{\textbf{1}} & \textbf{2} \\ \hline
Ideal           & \multicolumn{1}{c|}{27.968s}     & \multicolumn{1}{c|}{30.432s}     & 32.884s     \\ \hline
Proposed        & \multicolumn{1}{c|}{40.158s}     & \multicolumn{1}{c|}{48.63s}      & 53.097s     \\ \hline
Flock           & \multicolumn{1}{c|}{113.4s}      & \multicolumn{1}{c|}{156.79s}     & 179.31s     \\ \hline
\end{tabular}%
}
\vspace{0pt}
\end{table}
As shown in Fig.~\ref{fig:comparisonBoxPlot} and in Table~\ref{tab:sims}, the proposed method outperforms the flock method in all scenarios. Additionally, the average coverage time for the proposed method is similar to the ideal method, even with the communication limitation for the simulated failure scenarios. 

\section{Experiments}

\label{sec:exp}
The proposed approach was also validated through laboratory experiments with a multi-robot team. 
The team consists of one to two Husarion ROSbot 2.0 UGVs and one Bitcraze Crazyflie 2.1 using a Vicon motion capture system. 

The experiments effectively demonstrate all parts of the proposed approach, including intentional disconnections, coverage, gossiping, and task completion behaviors. In all experiments, the vehicles start within the communication range and are tasked to cover the environment and complete any discovered tasks. 

\begin{figure*}[t]
\centering
    \subfigure[]{
    \includegraphics[width = 0.155\textwidth,trim = {0cm 0cm 0cm 0},clip]{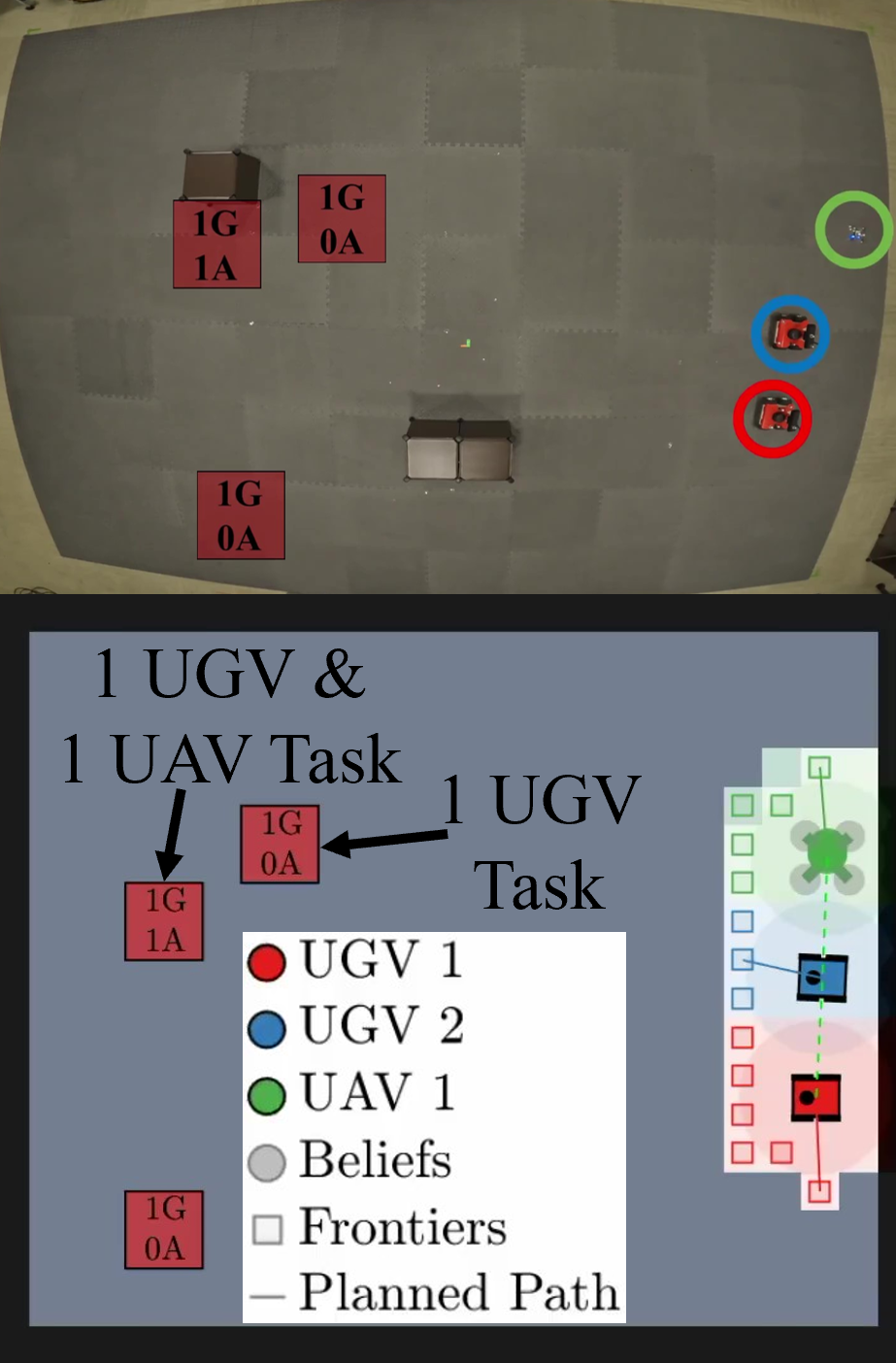}
    \label{fig:exa}
    }%
    \subfigure[]{
    \includegraphics[width = 0.155\textwidth,trim = {0cm 0cm 0cm 0},clip]{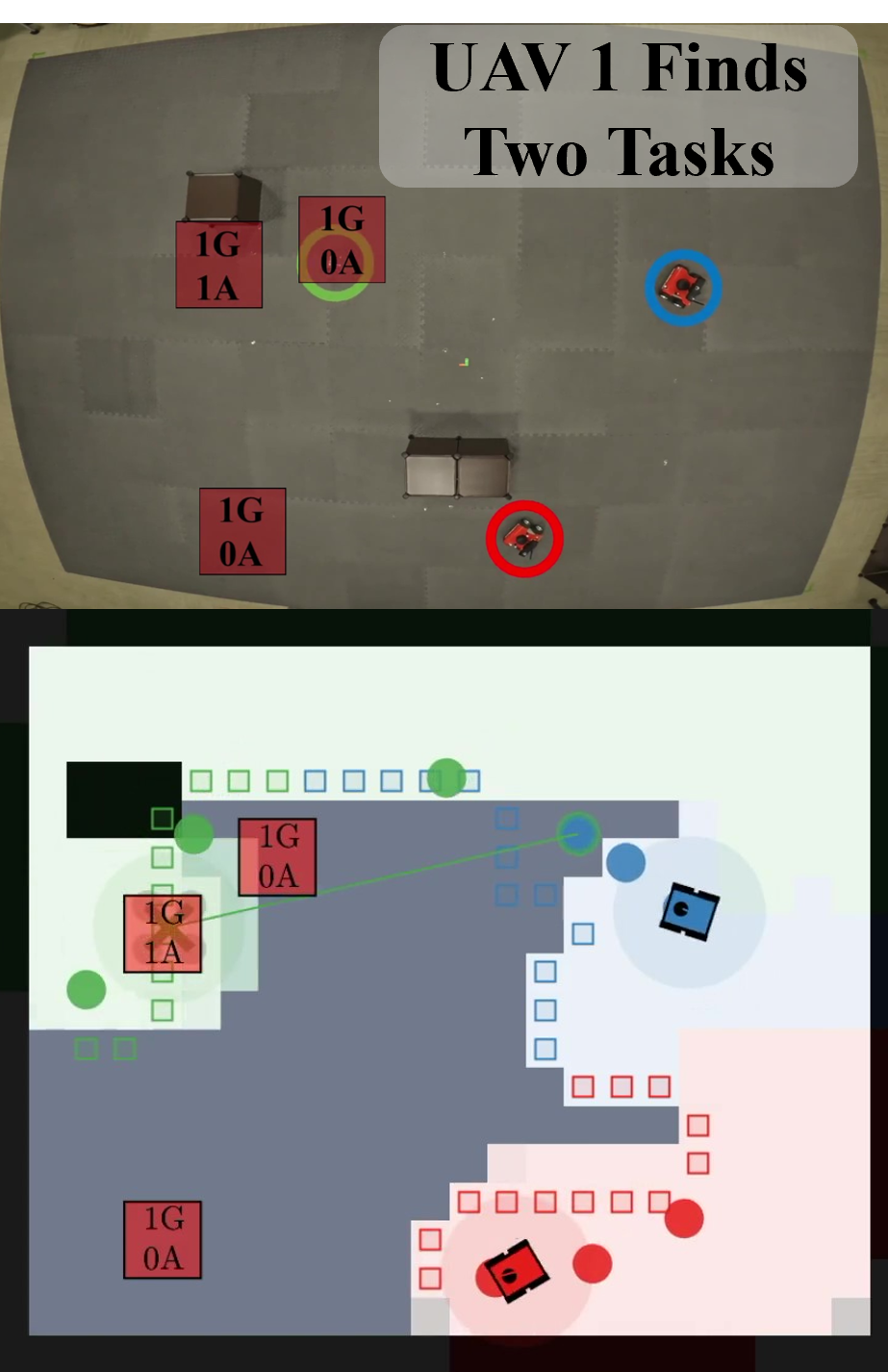}
    \label{fig:exb}
    }%
    \subfigure[]{
    \includegraphics[width = 0.155\textwidth,trim = {0cm 0cm 0cm 0},clip]{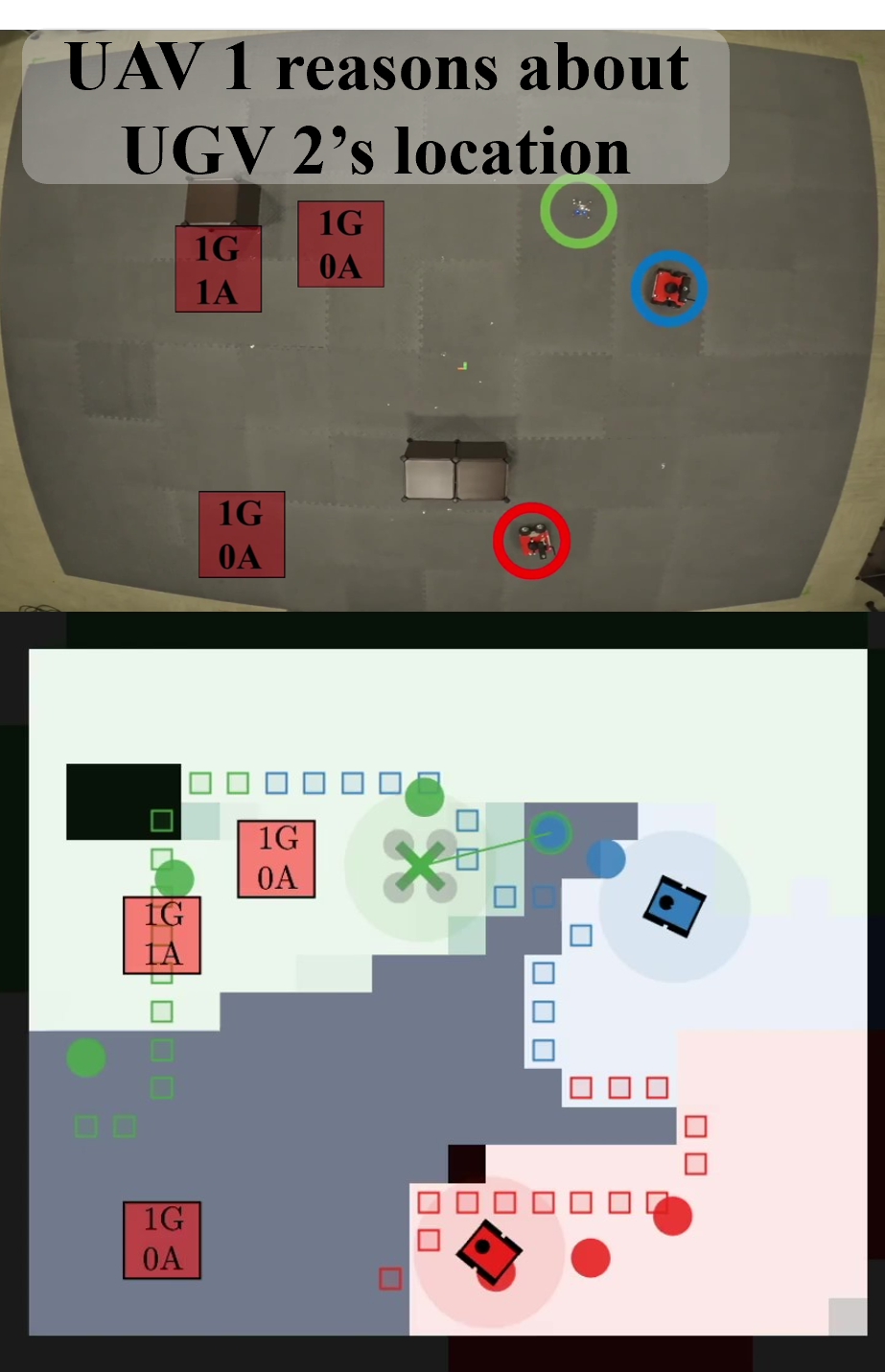}
    \label{fig:exc}
    }%
    \subfigure[]{
    \includegraphics[width = 0.155\textwidth,trim = {0cm 0cm 0cm 0},clip]{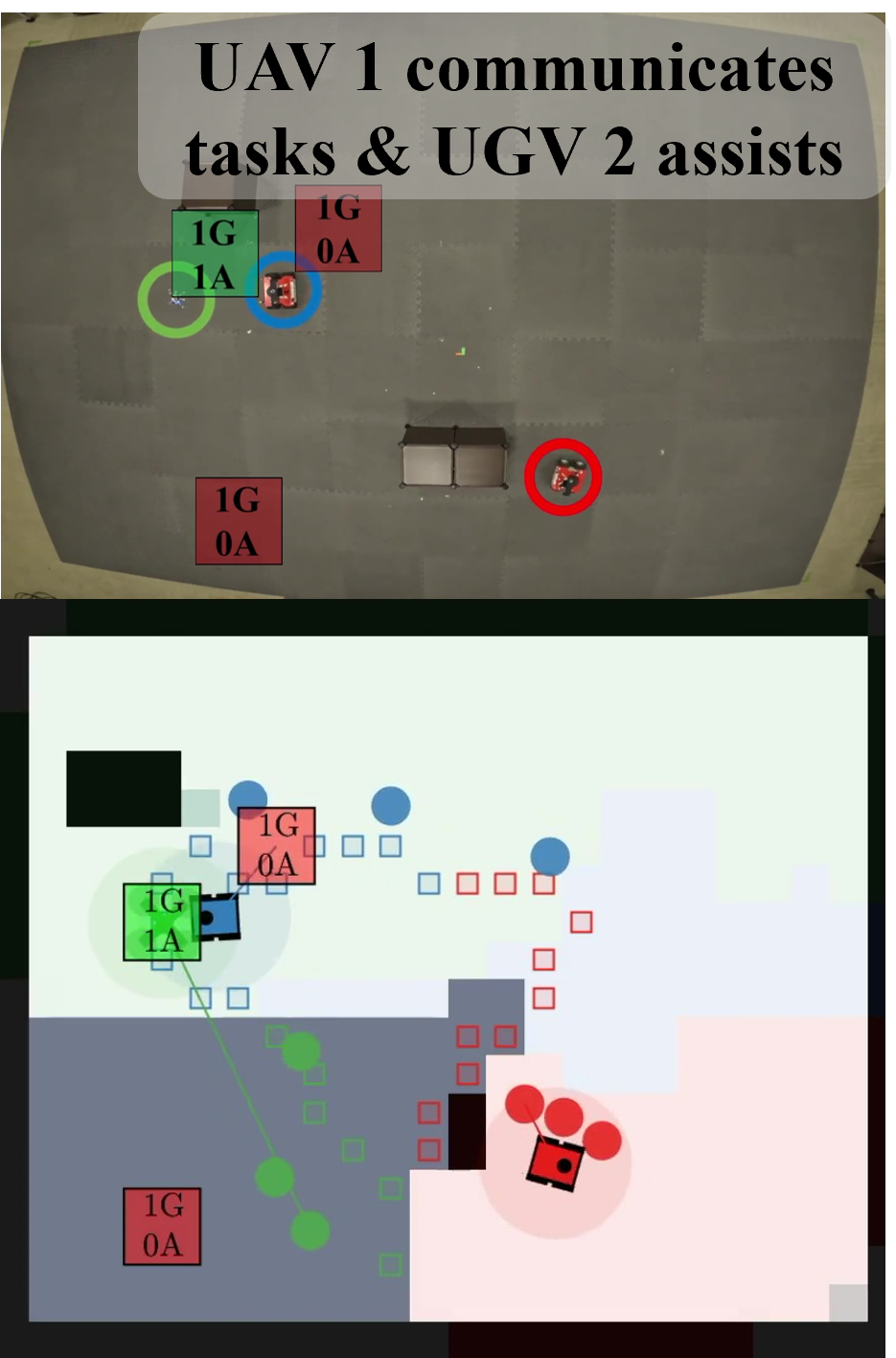}
    \label{fig:exd}
    }%
    \subfigure[]{
    \includegraphics[width = 0.155\textwidth,trim = {0cm 0cm 0cm 0},clip]{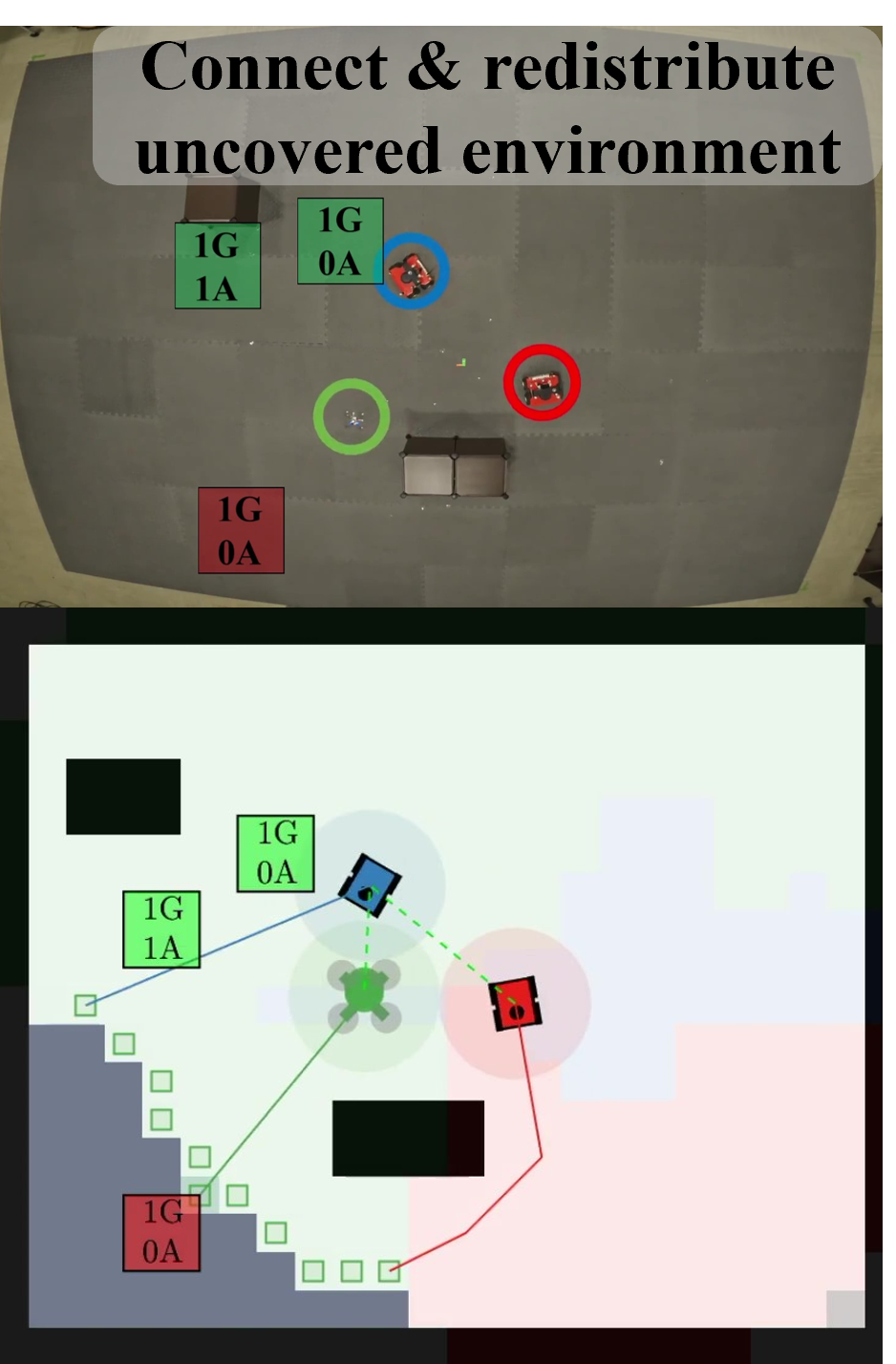}
    \label{fig:exe}
    }%
    \subfigure[]{
    \includegraphics[width = 0.155\textwidth,trim = {0cm 0cm 0cm 0},clip]{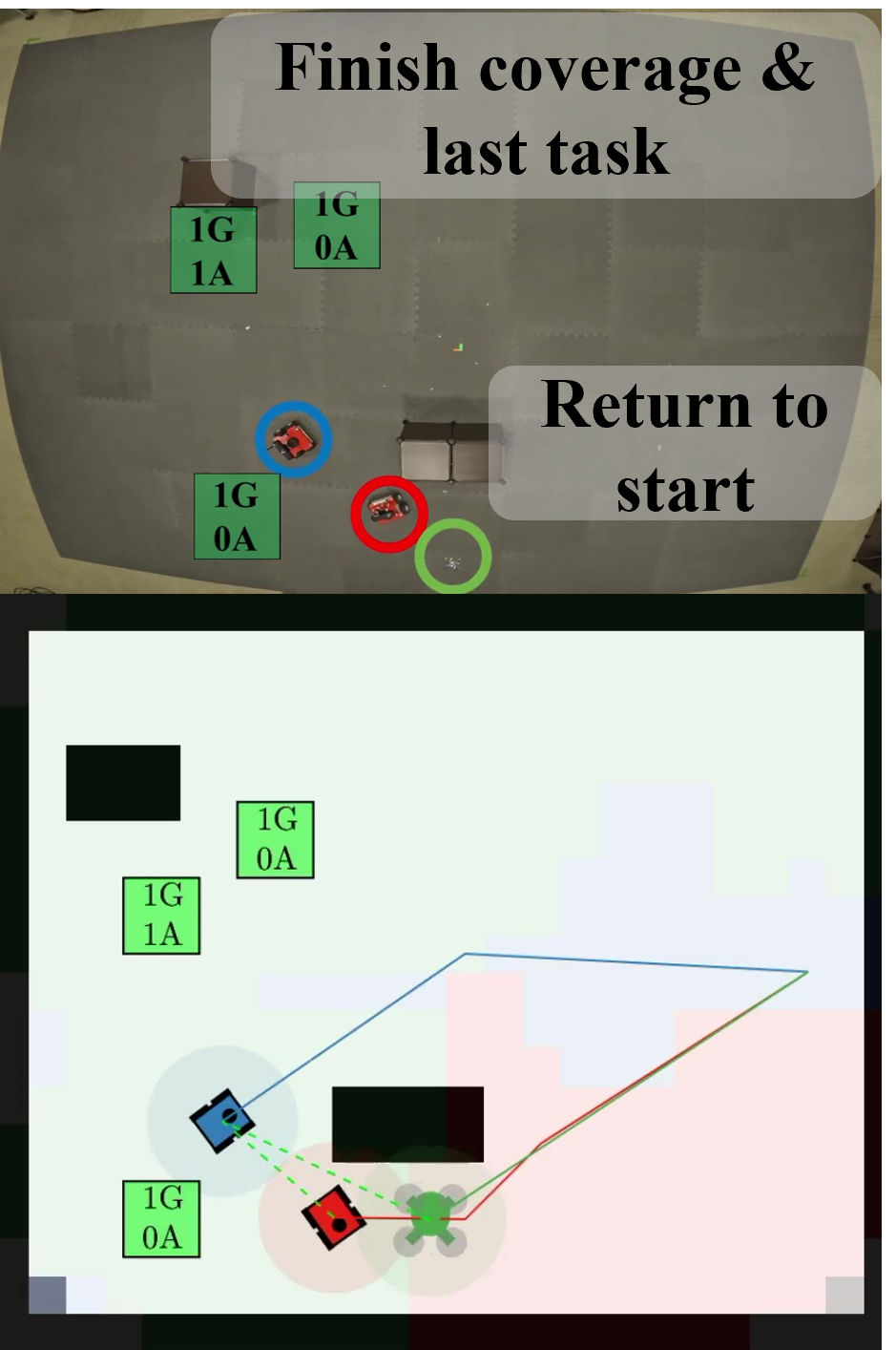}
    \label{fig:exf}
    }%
\vspace{-5pt}
\caption{Snapshots and results of an experimental case study.}
\label{fig:expFig}
\vspace{-15pt}
\end{figure*}

Experiments were performed in a $4$m$\times 5.5$m space containing convex obstacles and using, as a proof of concept, a sensing and communication range for each robot of $0.5$m. In Fig.~\ref{fig:expFig}, we show the results of a sample experiment where there are three tasks in the environment, but the total number of tasks and their locations are unknown. Two tasks require one ground vehicle, and one task requires an aerial and a ground vehicle simultaneously.

As shown in the figure, each robot is assigned a partition of the frontier in Fig.~\ref{fig:exa} that encourages the robots to disconnect. Once disconnected, UAV 1 finds a ground and aerial vehicle task and a ground vehicle task. UAV 1 plans to communicate these tasks with UGV 2's common belief or first particle, but UGV 2 has experienced a fault and is now tracking the third particle. Fig.~\ref{fig:exc} shows UAV 1 reasoning about the location of UGV 2 by iterating through its beliefs before finding UGV 2 at the third particle and communicating the tasks. In Fig.~\ref{fig:exd}, UGV 2 and UAV~1 complete the simultaneous task and UGV 2 plans to go to the ground vehicle task while UAV 1 plans to return to its common belief particle. In Fig.~\ref{fig:exe}, all vehicles' common belief particles converge to a common meeting place from \eqref{eq:commonMeeting}
and plan to cover the remaining environment and completing the last task before returning to their initial position in Fig.~\ref{fig:exf}. 
\section{Conclusion \& Future Work} \label{sec:concs}
In this work, we have presented a novel framework for heterogeneous multi-robot systems to leverage epistemic planning in its complex task allocation procedure while disconnected. The proposed method allows a heterogeneous MRS to disconnect and cooperatively plan according to a set of belief and empathy states. The generalized task allocation algorithm utilizes these belief states in allocating tasks while accounting for the potential need to gossip assignments to disconnected robots, which enables dynamic task allocation to be performed without the need for constant communication. We showcase the performance of our method compared to ideal communication and the flock method and apply our framework to real-world experiments.

From this, future theoretical work includes addressing the challenges of improving strategies for additional considerations such as complex obstacles in fully unknown environments. 
Furthermore, we would like to decrease the computation time for task allocation and optimize the necessary belief propagation for a larger multi-robot system by dividing the team into sub-teams. Outdoor experiments with our proposed implementation are also on the agenda. 
\section{Acknowledgement}
This work is based on research sponsored by Northrop Grumman through the University Basic Research Program and DARPA under Contract No. FA8750-18-C-0090. The authors also thank Shijie Gao for assisting with the experiments.


\bibliographystyle{IEEEtran}
\bibliography{newRef.bib}

\end{document}